# Dynamical prediction of two meteorological factors using the deep neural network and the long short-term memory (II)

Ki-Hong Shin[a], Jae-Won Jung[b], Ki-Ho Chang[c], Dong-In Lee[d, *], Cheol-Hwan You[d], Kyungsik Kim[a,b, *]

[a]*Department of Physics, Pukyong National University, Busan 48513, Republic of Korea*
[b]*DigiQuay Company Ltd., Seocho-gu Seoul 06552, Republic of Korea*
[c]*National Institute of Meteorological Research, Korea Meteorological Administration, Seogwipo 63568, Republic of Korea*
[d]*Department of Environmental Atmospheric Sciences, Pukyong National University, Busan 48513, Republic of Korea*

**Abstract**

This paper presents the predictive accuracy using two-variate meteorological factors, average temperature and average humidity, in neural network algorithms. We analyze result in five learning architectures such as the traditional artificial neural network, deep neural network, and extreme learning machine, long short-term memory, and long-short-term memory with peephole connections, after manipulating the computer-simulation. Our neural network modes are trained on the daily time-series dataset during seven years (from 2014 to 2020). From the trained results for 2500, 5000, and 7500 epochs, we obtain the predicted accuracies of the meteorological factors produced from outputs in ten metropolitan cities (Seoul, Daejeon, Daegu, Busan, Incheon, Gwangju, Pohang, Mokpo, Tongyeong, and Jeonju). The error statistics is found from the result of outputs, and we compare these values to each other after the manipulation of five neural networks. As using the long-short-term memory model in testing 1 (the average temperature predicted from the input layer with six input nodes), Tonyeong has the lowest root mean squared error (RMSE) value of 0.866 (%) in summer from the computer-simulation in order to predict the temperature. To predict the humidity, the RMSE is shown the lowest value of 5.732 (%), when using the long short-term memory model in summer in Mokpo in testing 2 (the average humidity predicted from the input layer with six input nodes). Particularly, the long short-term memory model is is found to be more accurate in forecasting daily levels than other neural network models in temperature and humidity forecastings. Our result may provide a computer-simuation basis for the necessity of exploring and develping a novel neural network evaluation method in the future.

---

* Corresponding authors. Fax: +82 51 629 4547.
E-mail addresses: kskim@pknu.ac.kr (K. Kim), leedi@pknu.ac.kr (D.-I. Lee).

# 1. Introduction

Recently, the meteorological factors including the wind speed, temperature, humidity, air pressure, global radiation, and diffuse radiation have been considerably concerned the climate variations for complex systems [1,2]. The statistical quantities of heat transfer, solar radiation, surface hydrology, and land subsidence [3] have been calculated within each grid cell of our earth with the weather prediction of the world meteorological organization (WMO), and these interactions are presently proceeding to be calculated to shed light on the atmospheric properties. Particularly, El Niño–southern oscillation (ENSO) forecast models have been categorized into three types: coupled physical models, statistical models, and hybrid models [4]. Among these models, the statistical models introduced for the ENSO forecasts have been the neural network model, multiple regression model, and canonical correlation analysis [5]. Barnston et al. [6] have found that the statistical models have reasonable accuracies in forecasting sea surface temperature anomalies. Recently, the machine learning has been considerable attention in the natural science fields such as statistical physics, particle physics, condensed matter physics, cosmology [7]. The research of statistical quantities on the spin models has particularly been simulated and analyzed in the restricted Boltzmann machine, restricted brief network, and recurrent neural network and so on [8,9].

The artificial intelligence has actively been applied in various fields and its research is progressed and developed. The neural network algorithm is a research method to optimize the weight of each node in all network layers in order to obtain a good prediction of output value. The past tick-data of scientific factors, as well-known, have made it difficult to predict the future situation combined with several factors. As paying attention to the developing potential of neural network algorithms, several models for the long short-term memory (LSTM) and the deep neural network (DNN), which are under study, are currently very successful from nonlinear and chaotic data in artificial intelligence fields.

The machine learning, which was once in a recession, has been applied to all fields as the era of big data has entered the era and applied to various industries and has established itself as a core technology. The machine learning improved its performance through learning, which includes supervised learning and unsupervised learning. Supervised learning used data with targets as input values, and unsupervised learning also used input data without targets [10]. Supervised learning includes regressions such as the linear regression, logistic regression, ridge regression, and Lasso regression, and classifications such as the support vector machine and the decision tree [11,12]. For the unsupervised learning, there are techniques such as the principle component analysis, K-means clustering, and density based spatial clustering of applications with Noise [13-15]. The reinforcement learning exists in addition to supervised and unsupervised learning. This learning method is known as a learning with actions and rewards. The AlphaGo has for example become famous for its against humans [16,17].

Over past eight decades, the neural network model have been proposed by McCulloch and Pitts [18]. Rosenblet [19] proposed the perceptron model, and the learning rules were firstly proposed by Hebb [20]. Minsky and Papert [21] have particularly advocated that perceptron is considered as a linear classifier that cannot solve the XOR problem. In the field of neural network, Rumelhart proposed a multilayer perceptron that added a hidden layer between the input layer and the output layer, and solved the XOR problem, and again faced the moment of development [22]. Until now, many models have been proposed for human memory as a collective property of neural networks. The neural network models introduced by Little [23] and Hopfield [24,25] have been based on an Ising Hamiltonian extended by equilibrium statistical mechanics. A detailed discussion of the equilibrium properties of the Hopfield model was discussed in Amit et al. [26,27]. Furthermore, Werbos has proposed backpropagation for learning the artificial neural network (ANN) [28], which was developed by Rumelhart in 1986, The backpropagation is a method of learning a neural network by calculating the error between the output value of the output layer calculated in the forward direction and the actual value propagated the error in the reverse direction. The backpropagation algorith is a delta rule and gradient descent method to update weights by performing learning in the direction of minimizing errors [29,30].

Indeed, Elman proposed a simple recurrent network using the output value of the hidden layer as the input value of the next time considering time [31]. The long short-term memory (LSTM) model, a variant of recurrent neural network that controls information flow by adding a gate to a node, was developed by Hochreiter and Schmidhuber [32]. The The long short-term memory with peephole connections (LSTM-PC) and the LSTM-GRU were developed from the LSTM [33,34]. In addition, the convolution neural network (CNN) used for high-level problems, including image recognition, object detection and language processing has been faced a new revival by Lecun et al. [35]. Furthermore, Huang et al. [36] proposed an extreme learning machine (ELM) to improve the slow progression of gradient descent-based algorithms due to iterative learning. The ELM is a single-hidden layer feedforward neural network with one hidden layer, without training, and uses a matrix to obtain the output value.

DNN is a special family of NNs described by multiple hidden layers between the input and output layers. This characterizes more complicated framework than traditional neural network, providing for exceptional capacity to learn a powerful feature representation from a big data. Deep neural networks have not been so far studied for some applied modeling, to our knowledge. DNN has various hyper-parameters such as the learning rate, drop out, epochs, batch size, hidden nodes, activation function and so on. In the case of weights, Xavier et al. [37–40] suggested that an initial weight value is set according to the number of nodes. In addition to the stochastic gradient descent (SGD), optimizers for the optimization such as the momentum, Nestrov, AdaGrad, RMSProp, Adam, and AdamW have also been developed [41–43]. Indeed, DNN continued to develop in several scientfic fields is applied to the other fields of stock market [44-52], transportation [53-60], weather [61-72], voice recognition [73-76], and electricity [77-83]. Tao et al. [84] have studied a state-of-the-art DNN for precipitation estimation using the satellite information, infrared, and water vapor channels, and they have particularly showed a two-stage framework for precipitation estimation from bispectral information. Although the stock market is a random and unpredictable field, DNN techniques are applied to predict the stock market [85,86]. The prediction accuracy was calculated by dividing the small, medium, large scale by applying deep learning with autoencoder and restricted Boltzmann machine, neural network with backpropagation algorithm, extreme learning machine (ELM), and radial basis function neural network [87]. Sermpinis et al. applied traditional statistical prediction techniques and ANN, RNN, and psi-sigma neural network for the EUR/USD exchange rate. Their results showed that the RMSE was smaller when the neural network model was used [88]. Vijh et al. [89] predicted the closing price of US firms using a single hidden layer neural network and a random forest model. Wang et al. applied the backpropagation neural network, Elman recurrent neural network, stochastic time effective neural network, and stochastic time effective function for SSE, TWSE, KOSPI, and Nikkei225. Authors have shown that artificial neural networks perform well in predicting the stock market [90].

Furthermore, Moustra et al. have introduced an ANN model to predict the intensity of earthquakes in Greece. They have used a multilayer perceptron for both seismic intensity time series data and seismic electric signals as input data [91]. Gonzalez et al. used the recurrent neural network and LSTM models to predict the earthquake intensity in Italy with hourly-data [92]. Kashiwao et al. predicted rain-autumn for the local regions in Japan. Authors were applied the hybrid algorithm in the random optimization method [93]. Zhang and Dong have studied the CNN model to predict the temperature by using the daily temperature data of China from 1952 to 2018 as learning data [94]. Bilgile et al. has used the ANN model to predict the temperature and precipitation in Turkey, and they simulated and analyzed 32 nodes with one hidden layer in this model. Their results have also showed a high correlation between the predicted value and the actual value [95]. Mohammadi et al. have collected weather data from Bandar Abass and Tabass with different weather conditions. Authors have predicted and compared daily dew point temperaure using the ELM, ANN, and SVM [96]. Maqsood et al. [97] have predicted the temperature, windspeed, and humidity by applying the multilayer perceptron, recurrent neural network, radial based function, and Hopfield model dring four seasons of Regina airport. Miao et al. [98] have developed a DNN composed of a convolution and LSTM recurrent module to estimate the predictive precipitation based on atmospheric dynamical fields. They compared the proposed model against the general circulation models and classical downscaling methods among several meteorological models.

Recently, in stock markets, Wei's work [99] has focused to the neural network models for stock price prediction using data selected from 20 stock datasets included the Shanghai and Shenzhen stock market in China. This was implemented six deep neural networks to predict stock prices and used four prediction error measures for evaluation. The results were also obtained that the prediction error value partially reflects the model accuracy of the stock price prediction, and cannot reflect the change in the direction of the model predicted stock price. Meanwhile, Mehtab et al. [100] have presented a suite of regression models using stock price data of a well-known company listed in the National Stock Exchange (NSE) of India during most recently two years. They showed from the experimental results that all of them yielded a high level of accuracy in their forecasting results, while the models exhibited wide divergence in predictive accuracies and execution speeds.

Indeed, several studies have existed in the published literature to predict the factors in meteorological community. Since the data properties of factors have intrinsically non-stationary, non-linear, and chaotic features, the predictive accuracy is regarded as a challenging task of the meteorological or climatological time-series prediction process. There is not uniquely sensitive to the suddenly unexpected change of meteorology and climate, similar to other scientific communities indicated time-series indices, but accurate predictions of meteorological time-series data are very precious for improving effective evolution strategies. The ANN, LSTM, and DNN methods have been successfully used for modelling and predicting time-series factor until now [101,102].

Due to combined complex interactions between meteorological factors, the predictive computations of temperature and humidity via computer simulation modeling of neural network is very difficult, because the past data analysis for existing neural network models play a crucial role for predictive accuracies. As well-known, there exist currently no more predictive studies of combined meteorological factors with high- and low-frequency data, and so high frequency data are due to be turned to the next time. Admittedly, the low-frequency data will be focused and interested in the research of five neural network models. In this paper, our objective is to study and analyze dynamical prediction of metrological factors (average temperature and humidity) using the neural network models in this paper. To benchmark the neural network models such as the ANN, DNN, ELM, LSTM, and LSTM-PC, we apply these to predict time-series data from ten different locations in Korea, namely Seoul, Daejeon, Daegu, Busan, Incheon, Gwangju, Pohang, Mokpo, Tongyeong, and Jeonju. This paper is organized as follows. Section 2 provides the brief formulas to the five neural network models for prediction accuracy. Corresponding calculations and the results for four statistical quantities, i.e., the root mean squared error, the mean absolute percentage error, the mean absolute error, and Theil's-U are presented in Section 3. The conclusions are presented and future studies are discussed in Section 4.

## 2. Methodology

In this section, we simply recall the method and its technique for five NN models, that is, the artificial neural network (ANN), Deep Neural network (DNN), and extreme learning machine (ELM), long short-term memory (LSTM), and long short-term memory with peephole connections (LSTM-PC).

**2-1. Artificial neural network (ANN), deep neural network (DNN), and extreme machine learning (EML)**

ANN is a mathematical model that presents some features of brain functions as a computer-simulation. That is, it is an artificially explored network, distinguished from a biological neural network. The basic structure of the ANN has three layers: input, hidden, and output. Each layer is determined by connection weight and its bias. In an arbitrary layer of the neural network, each node constitutes as one neuron, and one link between nodes means one connection weight of a synapse. The connection weight is corrected as feedback via a training phase, and is designed to implement self-learning.

The neural network architectures are set up to compare the ANN performance. When constructing the ANN model, data variables should be normalized to the interval between 0 and 1 as follows: $x_{\text{normal}} = (x - x_{\min})/(x_{\max} - x_{\min})$. The ANN structure is shown as follows: In input layer, $T_{t-2}$, and $T_{t-1}$, $T_t$ denote the input nodes for temperature, and $H_{t-2}$, $H_{t-1}$, and $H_t$ the input nodes for humidity at time lag $t-2$, $t-1$, and $t$. $HL_1$, $HL_2$, and $HL_3$ are the hidden nodes in the hidden layer. $T_{t+1}$ ($H_{t+1}$) denotes the output node of temperature (humidity) prediction at time lag $t+1$. We can find the optimal patterns of the output after learning iteratively. In our study, we limit and construct a DNN as one input with four nodes and two hidden layers with each three nodes, as shown in Fig. 1. That is, from the input layer, $T_{t-1}$ ($T_t$) denotes the input node for temperature at time lag $t-1(t)$, and $H_{t-1}$ ($H_t$) the input node for humidity at time lag $t-1(t)$. In first (second) hidden, layer, $HL^1_1$, $HL^1_2$, and $HL^1_3$ ($HL^2_1$, $HL^2_2$, and $HL^2_3$) are the hidden nodes. $T_{t+1}$ ($H_{t+1}$) denotes the output node of temperature (humidity) prediction at time lag $t+1$.

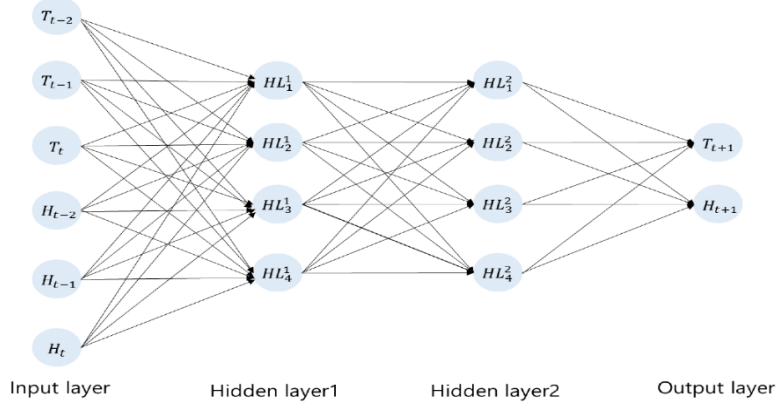

Fig. 1: DNN framework with one input and two hidden layers.

The calculation of ANN and DNN for the weighted sum of an external stimulus entered into the input layer can be produced the output as the pertinent reaction through the activation function. That is, the weighted sum of the external stimulus is represented in terms of $y_j = \sum_{i=1}^{n} w_{ij} x_i$, where $x_i$ is the external stimulus, and $w_{ij}$ the connection weight of output neuron. As the reaction value of the output neuron is determined by the activation function, we use the sigmoid function $\sigma(y_j) = 1/1 + \exp(-\alpha y_j)$ for the analog output. Here, we generally use $\alpha=1$ in order to avoid the divergence of $\sigma(y_j)$ and to obtain the optimal value of the connection weight of output. The learning of the counter propagation algorithm can minimize the sum of errors, as compared with the calculated value of all directional feed forward to the target value.

ELM is a feedforward neural network for classification, regression, and clustering. Let us recall the ELM model [74,75], and this is a feature learning with a single layer of hidden nodes. Using a set of training samples $[(x_j, y_j)]_{j=1}^{s}$ for $s$ samples and $v$ classes, the activation function $g_i(x_j)$ for the single hidden layer with $n$ nodes is given by

$$z_j = \sum_{i=1}^{n} \alpha_i g_i(x_j) = \sum_{i=1}^{n} \alpha_i g_i(w_i \cdot x_j + b_i), \; j = 1,2,\ldots,s. \tag{1}$$

Here, $x_j = [x_{j1}, x_{j2}, \ldots, x_{js}]^T$ is the input units, and $y_j = [y_{j1}, y_{j2}, \ldots, y_{jv}]^T$ the output units. The statistical quantity $w_i = [w_{j1}, w_{j2}, \ldots, w_{js}]^T$ denotes the connecting weights of hidden unit $i$ to input units, $b_i$ the bias of hidden unit $i$, $\alpha_i = [\alpha_{i1}, \alpha_{i2}, \ldots, \alpha_{iv}]^T$ the connecting weights of hidden unit $i$ to the output units, and $z_j$ the actual network output. We can our ELM model can solve using error minimization as $\min_\beta \|G\alpha - C\|_f$ with

$$G(w_1, \ldots, w_n, b_1, \ldots, b_n) = \begin{bmatrix} g(w_1 \cdot x_1 + b_1) & \cdots & g(w_{\tilde{N}} \cdot x_1 + b_n) \\ \vdots & \cdots & \vdots \\ g(w_1 \cdot x_s + b_1) & \cdots & g(w_{\tilde{N}} \cdot x_s + b_n) \end{bmatrix}_{n \times s} \tag{2}$$

and

$$C = [C_1^T, C_2^T, \ldots, C_s^T]^T. \tag{3}$$

Here, $G(\mathbf{w}, \mathbf{b})$ is the output matrix of hidden layer, $\alpha$ the output weight matrix, and $C$ the output matrix in Eq. (3). The ELM randomly selects the hidden unit parameters, and the output weight parameters need to be determined.

**2-2. Long short-term memory (LSTM) and long short-term memory with peephole connections (LSTM-PC)**

The recurrent neural network is known as a distinctive and general algorithm for processing long-time series data, and this is possible to learn by using the novel input data from subsequent step and the output data from the previous step at the same time. The LSTM is known to be advantageous both for preventing the inherent vanishing gradient problem of general recurrent neural network and for predicting time series data [81,82].

The LSTM has previously introduced as a novel type of recurrent neural network, and this is considered the better model than the recurrent neural network on tasks involving long time lags. The architecture of LSTM have played a crucial role in connecting long time lags between input events. Let us recall the LSTM is composed of a cell with three gates attached as follows. As well-known, the LSTM cell that has the ability to bridge very long time lags is divided into forget, input, and output gates to protect and control the cell state. Let us recall one cell of LSTM. As the first step in the cell, the forget gate $f_t$ (input gate $i_t$) enters the input value $x_t$ and an output value $h_{t-1}$ through the previous step, where $x_t$ and $h_{t-1}$ are the normalized values between zero and one. If the output information is exited, $f_t$ is represented in terms of

$$i_t = \sigma(w_{xi}x_t + w_{hi}h_{t-1} + b_i), \tag{4}$$
$$f_t = \sigma(w_{xf}x_t + w_{hf}h_{t-1} + b_f). \tag{5}$$
$$\overline{c_t} = \tanh(w_{xc}x_t + w_{hc}h_{t-1} + b_c), \tag{6}$$
$$c_t = f_t \odot c_{t-1} + i_t \odot \overline{c_t}. \tag{7}$$

Here, $\sigma(y)$ denotes the activation function as a function of y, $w_{xi}$ and $w_{xi}$ the weights of gate, and $b_i$ and $b_f$ an bias value. In the second step, the input gate has the new cell states updated, when subsequent values are entered as follows:

$$\overline{c_t} = \tanh(w_{xc}x_t + w_{hc}h_{t-1} + b_c), \tag{6}$$
$$c_t = f_t \odot c_{t-1} + i_t \odot \overline{c_t}. \tag{7}$$
$$o_t = \sigma(w_{xo}x_t + w_{ho}h_{t-1} + b_o). \tag{8}$$

Here, $c_t$ is the cell state, and $\overline{c_t}$ the activation function created through the output gate. $w_{xc}$ and $w_{hc}$ are the weight values in output gate, and $b_c$ the bias value. The symbol $\odot$ represents the inner product of matrices. The output gate $o_t$ is exited in the third step. Lastly, an new predicted output value $h_t$ is calculated as $h_t = o_t \odot \tanh c_t$.

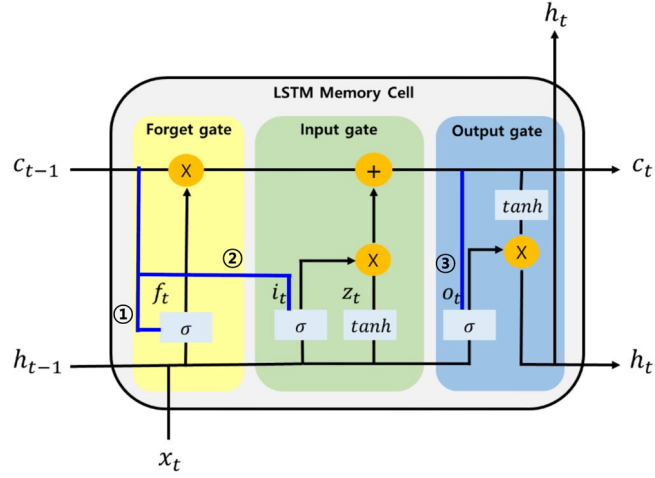

**Fig. 2:** LSTM-PC structure, where the blue paths ①, ②, and ③ are the peephole connections connected from $c_{t-1}$ and $c_t$, respectively.

The recurrent neural network has in principle learned to make use of numerous sequential tasks such as motor control and rhythm detection. It is well known that the LSTM outperforms other recurrent neural networks on tasks involving long time lags. We find that long short-term memory with peephole connections (LSTM-PC) augmented from the internal cells to the multiplicative gates can learn the fine distinction between sequences of spikes [79]. This model constitutes the same LSTM cell as the forget gate, input gate, and output gate.

The peephole connections allow the gates to carry out their operations as a function of both the incoming inputs and the previous state of the cell. In Fig. 2, the LSTM implements the compound recursive function to obtain the predicted output value $h_t$ as follows:

$$i_t = \sigma(w_{xi}x_t + w_{hi}h_{t-1} + w_{ci}c_{t-1} + b_i) \tag{9}$$

$$f_t = \sigma(w_{xf}x_t + w_{hf}h_{t-1} + w_{cf}c_{t-1} + b_f) \tag{10}$$

$$c_t = f_t \odot c_{t-1} + i_t \odot \tanh(w_{xc}x_t + w_{hc}h_{t-1} + b_c) \tag{11}$$

$$o_t = \sigma(w_{xo}x_t + w_{ho}h_{t-1} + w_{co}c_t + b_o) \tag{12}$$

$$h_t = o_t \odot \tanh c_t \,. \tag{13}$$

where $i_t$, $f_t$, $c_t$, and $o_t$ are the input gate, forget gate, cell gate, and output gate activation vectors at time lag $t$, respectively, and $w_{ci}$, $w_{cf}$, and $w_{co}$ the peephole weights. In view of Eqs. (9), (10), and (12), we can discriminate the LSTM-PC from the LSTM. From Fig. 2, the path ① (②) has the value of $w_{ci}c_{t-1}$ ($w_{cf}c_{t-1}$) added in one peephole connection connected from $c_{t-1}$ to forget gate (input gate). In path ③, $w_{co}c_t$ is the value added in one peephole connection connected from $c_t$ to output gate. It will be particularly showed in section 3 that the LSTM and LSTM-PC are set for different train set sizes over 2500, 5000, and 7500 epochs.

After we continuously adjust the connection weight for a neural network model, we iterate the learning process. Lastly, the predictive accuracies of neural network models are performed using the root mean squared error (RMSE), the mean absolute percentage error (MAPE), the mean absolute error (MAE), and Theil's-U statistics as follows:

$$\text{RMSE} = [\frac{1}{N}\sum_{i=1}^{N}(y_i - \overline{y_i})^2]^{1/2} \tag{14}$$

$$\text{MAPE} = \frac{1}{N}\sum_{i=1}^{N}\left|\frac{y_i - \overline{y_i}}{y_i}\right| \times 100\% \tag{15}$$

$$\text{MAE} = \frac{1}{N}\sum_{i=1}^{N}|y_i - \overline{y_i}| \tag{16}$$

$$\text{Theil's U} = \frac{[\frac{1}{N}\sum_{i=1}^{N}(y_i - \overline{y_i})^2]^{1/2}}{[\frac{1}{N}\sum_{i=1}^{N}y_i^2]^{1/2} + [\frac{1}{N}\sum_{i=1}^{N}P\overline{y}_i^2]^{1/2}} \ . \tag{17}$$

Here, $y_i$ and $\overline{y_i}$ represent an actual value and a predicted value, respectively. The RMSE is particularly the square root of the MSE. Since introducing the square root, we can make that the error range is the same as the actual value range. Indeed, the RMSE and the MSE are very similar, but they cannot be interchanged with each other for gradient-based methods.

## 3. Numerical results and predictions

**3-1. Data and testings**

In this study, as our data, we use the temperature and the humidity of ten cities in South Korea extracted from the Korea Meteorological Administration (KMA). The metropolitan ten cities we studied and analyzed are Seoul, Incheon, Daejeon, Daegu, Busan, Pohang, Tongyeong, Gwangju, Mokpo, and Jeonju. We extract the data of the manned regional meteorological offices of the KMA to ensure the reliability of data, and these are for seven years from 2014 to 2020. In this study, we use daily training data ~85%, from March 2014 to February 2019, to train the neural network models and remained testing data ~15% for prediction, from March 2019 to February 2020, to test the predictive accuracy of the methods for two meteorological factors (temperature and humidity). We also use and calculate data for the four seasons divided into the spring (March, April, May), the summer (June, July, August), the autumn (September, October, November), and the winter (December, January, February).

From Ref. [103], we have described the details of our computer-simulation and present the results obtained by testing 1 and 2. In the prediction case of temperature and humidity, we completed two different test cases as follows: That is, testing 1 of predictive value $T_{t+1}$ and testing 2 of predictive value $H_{t+1}$ denoted four input nodes $T_{t-1}, T_t, H_{t-1}$, and $H_t$. We tested the predicted accuracies for temperature $T_{t+1}$ and humidity $H_{t+1}$ at time lag $t+1$. In Appendix A, we previously performed for two testings. That is, testing A1(see Table A1 in Appendix A) has the four nodes $T_{t-1}, T_t, H_{t-1}, H_t$, in the input layer and the one output node $T_{t+1}$ in output layer, and testing A2 (see Table A2 in Appendix A) has also the four input nodes $T_{t-1}, T_t, H_{t-1}, H_t$, and the one output node $H_{t+1}$. We are due to compare to each other from our predictive result.

In this subsection, we introduce testing 3 of predictive value $T_{t+1}$ and testing 4 of predictive value $H_{t+1}$, denoted six input nodes $T_{t-2}, T_{t-1}, T_t, H_{t-2}, H_{t-1}, H_t$, in all neural network structures. Here, We select for most of the benchmarks when inceasing the learning rate (lr) both from 0.1 to 0.5 and from 0.001 to 0.009, similar to that of Ref. [103]. For testing 3 and 4, we set the five learning rates 0.1, 0.2, 0.3, 0.4, 0.5 for the ANN and the DNN, while the learning rates for LSTM and LSTM-PC are set as 0.001, 0.003, 0.005, 0.007, 0.009, for different train set sizes over three runs (2500, 5000, and 7500 epochs). The predicted values of the ELM are obtained by averaging the results over 2500, 5000, and 7500 epochs. A prediction model is created and the average of the prediction values is obtained through the prediction model was used as the final prediction value.

## 3-2. Numerical calculations

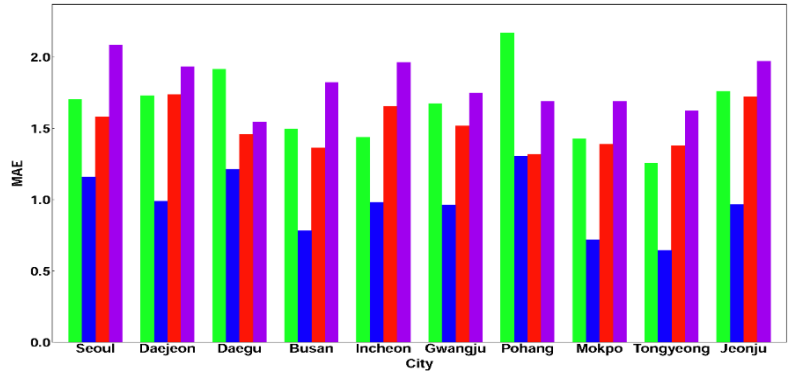

**Fig. 3:** The lowest MAE of five neural network models for all three kinds of epochs in spring (green bar), summer (blue bar), autumn (red bar), winter (purple bar) of testing 3.

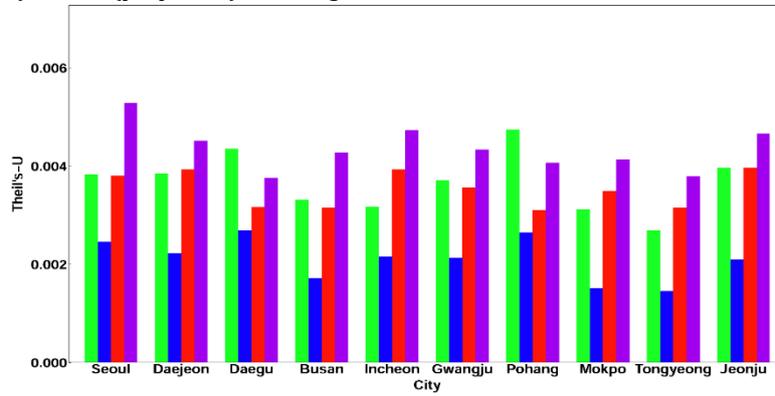

**Fig. 4:** The lowest Theil's-U of five neural network models for all three kinds of epochs in spring (green bar), summer (blue bar), autumn (red bar), winter (purple bar) of testing 3.

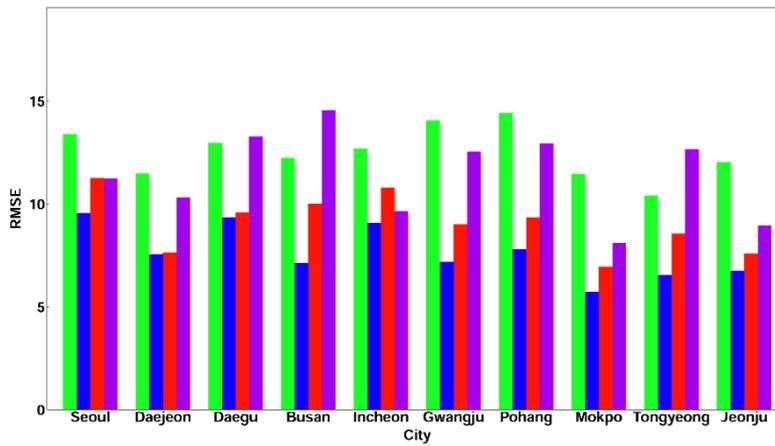

**Fig. 5:** The lowest RMSE of five neural network models for all three kinds of epochs in spring (green bar), summer (blue bar), autumn (red bar), winter (purple bar) of testing 4.

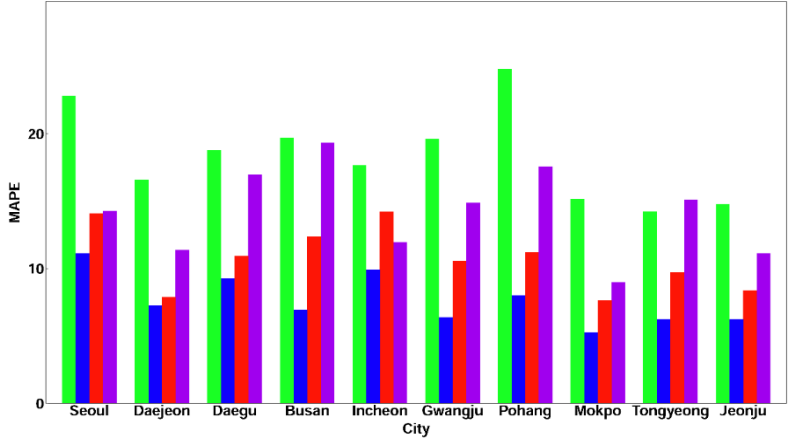

**Fig. 6:** The lowest MAPE of five NN models for all three kinds of epochs in spring (green bar), summer (blue bar), autumn (red bar), winter (purple bar) of testing 4.

**Table. 1:** The RMSE, MAPE, MAE and Theil's-U values (unit: %) in testing 3, where lr denotes the learning rate.

| City | Season | RMSE | MAPE | MAE | Theil-U ($\times 10^{-3}$) |
|---|---|---|---|---|---|
| Seoul | Spring | 2.19(DNN lr = 0.1) | 0.595(DNN lr = 0.1) | 1.703(DNN lr = 0.1) | 3.826(DNN lr = 0.1) |
| | Summer | 1.468(ANN lr = 0.3) | 0.389(ANN lr = 0.3) | 1.16(ANN lr = 0.3) | 2.46(ANN lr = 0.3) |
| | Autumn | 2.195(LSTM lr = 0.005) | 0.555(LSTM lr = 0.003) | 1.583(LSTM lr = 0.003) | 3.807(LSTM lr = 0.005) |
| | Winter | 2.901(DNN lr = 0.3) | 0.761(DNN lr = 0.3) | 2.087(DNN lr = 0.3) | 5.278(DNN lr = 0.3) |
| Daejeon | Spring | 2.207(ANN lr = 0.1) | 0.604(ANN lr = 0.1) | 1.732(ANN lr = 0.1) | 3.85(ANN lr = 0.1) |
| | Summer | 1.322(LSTM-PC lr = 0.001) | 0.333(LSTM lr = 0.003) | 0.99(LSTM lr = 0.003) | 2.216(LSTM-PC lr = 0.001) |
| | Autumn | 2.268(LSTM lr = 0.005) | 0.607(LSTM lr = 0.005) | 1.74(LSTM lr = 0.005) | 3.93(LSTM lr = 0.005) |
| | Winter | 2.492(ANN lr = 0.3) | 0.699(DNN lr = 0.3) | 1.932(DNN lr = 0.3) | 4.519(ANN lr = 0.3) |
| Daegu | Spring | 2.509(ANN lr = 0.1) | 0.666(ANN lr = 0.1) | 1.916(ANN lr = 0.1) | 4.361(ANN lr = 0.1) |
| | Summer | 1.605(LSTM lr = 0.003) | 0.408(ANN lr = 0.3) | 1.215(ANN lr = 0.3) | 2.688(LSTM lr = 0.003) |
| | Autumn | 1.832(LSTM lr = 0.005) | 0.508(LSTM lr = 0.001) | 1.461(LSTM lr = 0.001) | 3.168(LSTM lr = 0.005) |
| | Winter | 2.081(LSTM-PC lr = 0.009) | 0.559(DNN lr = 0.1) | 1.547(DNN lr = 0.1) | 3.756(LSTM-PC lr = 0.009) |
| Busan | Spring | 1.904(DNN lr = 0.1) | 0.521(DNN lr = 0.1) | 1.498(DNN lr = 0.1) | 3.31(DNN lr = 0.1) |
| | Summer | 1.023(LSTM-PC lr = 0.003) | 0.264(ANN lr = 0.1) | 0.782(ANN lr = 0.1) | 1.72(LSTM-PC lr = 0.003) |
| | Autumn | 1.832(LSTM-PC lr = 0.001) | 0.472(LSTM-PC lr = 0.001) | 1.365(LSTM-PC lr = 0.001) | 3.145(LSTM-PC lr = 0.1) |
| | Winter | 2.396(ANN lr = 0.3) | 0.652(ANN lr = 0.3) | 1.824(ANN lr = 0.3) | 4.279(ANN lr = 0.3) |
| Incheon | Spring | 1.814(ANN lr = 0.1) | 0.504(ANN lr = 0.1) | 1.437(ANN lr = 0.1) | 3.182(DNN lr = 0.1) |
| | Summer | 1.288(LSTM-PC lr = 0.009) | 0.33(ANN lr = 0.1) | 0.983(ANN lr = 0.1) | 2.163(LSTM-PC lr = 0.009) |
| | Autumn | 2.268(ANN lr = 0.3) | 0.581(LSTM-PC lr = 0.001) | 1.655(LSTM-PC lr = 0.001) | 3.926(ANN lr = 0.3) |
| | Winter | 2.604(DNN lr = 0.3) | 0.715(DNN lr = 0.3) | 1.963(DNN lr = 0.3) | 4.734(DNN lr = 0.3) |
| Gwangju | Spring | 2.127(LSTM | 0.586(LSTM | 1.676(LSTM | 3.705(LSTM |

| City | Season | RMSE | MAPE | MAE | Theils'-U ($10^{-3}$) |
|---|---|---|---|---|---|
| | | lr = 0.003) | lr = 0.005) | lr = 0.005) | lr = 0.003) |
| | Summer | 1.267(LSTM-PC lr = 0.001) | 0.325(LSTM lr = 0.001) | 0.965(LSTM lr = 0.001) | 2.126(LSTM-PC lr = 0.001) |
| | Autumn | 2.069(LSTM lr = 0.001) | 0.529(LSTM-PC lr = 0.5) | 1.518(LSTM-PC lr = 0.005) | 3.569(LST lr = 0.001) |
| | Winter | 2.412(LSTM lr = 0.003) | 0.629(DNN lr = 0.3) | 1.75(DNN lr = 0.3) | 4.343(ANN lr = 0.3) |
| Pohang | Spring | 2.737(LSTM lr = 0.001) | 0.753(LSTM lr = 0.001) | 2.171(LSTM lr = 0.001) | 4.752(LSTM lr = 0.001) |
| | Summer | 1.582(DNN lr = 0.1) | 0.439(LSTM lr = 0.001) | 1.306(LSTM lr = 0.001) | 2.654(DNN lr = 0.1) |
| | Autumn | 1.806(LSTM-PC lr = 0.001) | 0.459(LSTM-PC lr = 0.001) | 1.321(LSTM-PC lr = 0.001) | 3.107(LSTM-PC lr = 0.001) |
| | Winter | 2.27(ANN lr = 0.3) | 0.607(LSTM lr = 0.009) | 1.688(LSTM lr = 0.009) | 4.073(ANN lr = 0.3) |
| Mokpo | Spring | 1.785(LSTM lr = 0.003) | 0.243(LSTM lr = 0.003) | 1.426(LSTM lr = 0.003) | 3.12(LSTM lr = 0.003) |
| | Summer | 0.896(DNN lr = 0.1) | 0.243(ANN lr = 0.1) | 0.721(ANN lr = 0.1) | 1.506(DNN lr = 0.1) |
| | Autumn | 2.025(LSTM-PC lr = 0.005) | 0.485(LSTM-PC lr = 0.001) | 1.389(LSTM-PC lr = 0.001) | 3.496(LSTM-PC lr = 0.005) |
| | Winter | 2.294(DNN lr = 0.1) | 0.609(ANN lr = 0.3) | 1.689(ANN lr = 0.3) | 4.14(DNN lr = 0.1) |
| Tongyeong | Spring | 1.547(ANN lr = 0.1) | 0.438(LSTM lr = 0.007) | 1.256(LSTM lr = 0.007) | 2.695(ANN lr = 0.1) |
| | Summer | 0.866(LSTM lr = 0.005) | 0.218(LSTM lr = 0.005) | 0.647(LSTM lr = 0.005) | 1.457(LSTM lr = 0.005) |
| | Autumn | 1.832(LSTM-PC lr = 0.001) | 0.477(LSTM-PC lr = 0.001) | 1.38(LSTM-PC lr = 0.001) | 3.15(LSTM-PC lr = 0.001) |
| | Winter | 2.118(ANN lr = 0.3) | 0.581(DNN lr = 0.1) | 1.624(DNN lr = 0.1) | 3.791(ANN lr = 0.3) |
| Jeonju | Spring | 2.272(ANN lr = 0.1) | 0.616(ANN lr = 0.1) | 1.762(ANN lr = 0.1) | 3.968(ANN lr = 0.1) |
| | Summer | 1.25(ANN lr = 0.1) | 0.325(ANN lr= 0.1) | 0.967(ANN lr = 0.1) | 2.097(ANN lr = 0.1) |
| | Autumn | 2.296(ANN lr = 0.1) | 0.6(LSTM lr = 0.005) | 1.722(LSTM lr = 0.005) | 3.97(ANN lr = 0.1) |
| | Winter | 2.579(ANN lr = 0.3) | 0.712(ANN lr = 0.3) | 1.972(ANN lr = 0.3) | 4.658(ANN lr = 0.3) |

**Table. 2:** The RMSE, MAPE, MAE and Theil's-U values (unit: %) in testing 4, where lr denotes the learning rate.

| City | Season | RMSE | MAPE | MAE | Theils'-U ($10^{-3}$) |
|---|---|---|---|---|---|
| Seoul | Spring | 13.398 (LSTM lr = 0.001) | 22.798 (LSTM-PC lr = 0.003) | 10.6(LSTM-PC lr = 0.003) | 0.127(LSTM lr = 0.001) |
| | Summer | 9.575(LSTM-PC lr = 0.009) | 11.136(LSTM lr = 0.007) | 7.17(LSTM lr = 0.007) | 0.071(LSTM lr = 0.001) |
| | Autumn | 11.266(LSTM lr = 0.001) | 14.082(LSTM lr = 0.005) | 8.01(LSTM lr = 0.001) | 0.091(LSTM lr = 0.001) |
| | Winter | 11.232(ANN lr = 0.1) | 14.262(ANN lr = 0.1) | 8.377(ANN lr = 0.1) | 0.101(LSTM-PC lr = 0.003) |
| Daejeon | Spring | 11.497(LSTM-PC lr = 0.003) | 16.577(LSTM-PC lr = 0.003) | 9.536(LSTM-PC lr = 0.003) | 0.095(LSTM-PC lr = 0.003) |
| | Summer | 7.554(LSTM lr = 0.009) | 7.267(LSTM lr = 0.007) | 5.696(LSTM lr = 0.007) | 0.049(LSTM lr = 0.009) |
| | Autumn | 7.657(ANN lr = 0.001) | 7.898(ANN lr = 0.1) | 5.742(ANN lr = 0.1) | 0.051(ANN lr = 0.1) |
| | Winter | 10.32 (ELM) | 11.379(LSTM lr = 0.001) | 7.914(LSTM lr = 0.001) | 0.076 (ELM) |
| Daegu | Spring | 12.969(LSTM-PC lr = 0.001) | 18.761(LSTM-PC lr = 0.007) | 9.748(LSTM-PC lr = 0.007) | 0.123(LSTM-PC lr = 0.001) |
| | Summer | 9.354(LSTM lr = 0.009) | 9.275(LSTM-PC lr = 0.003) | 6.977(LSTM-PC lr = 0.003) | 0.065(LSTM lr = 0.009) |
| | Autumn | 9.602(LSTM-PC lr = 0.003) | 10.922(LSTM-PC lr = 0.003) | 7.506(LSTM lr = 0.001) | 0.068(LSTM-PC lr = 0.003) |
| | Winter | 13.274(LSTM-PC lr = 0.003) | 16.955(ANN lr = 0.1) | 9.534(LSTM-PC lr = 0.003) | 0.111(LSTM-PC lr = 0.003) |
| Busan | Spring | 12.227(LSTM | 19.691(LSTM | 10.047(LSTM | 0.101(LSTM |
| | | lr = 0.001) | lr = 0.001) | lr = 0.001) | lr = 0.001) |
| | Summer | 7.139(ANN lr = 0.1) | 6.928(ANN lr = 0.1) | 5.556(ANN lr = 0.1) | 0.045(ANN lr = 0.1) |
| | Autumn | 10.029(LSTM-PC lr = 0.001) | 12.362(LSTM-PC lr = 0.1) | 7.705(LSTM-PC lr = 0.001) | 0.072(LSTM-PC lr = 0.001) |
| | Winter | 14.565(ANN lr = 0.3) | 19.332(LSTM lr = 0.005) | 10.424(LSTM lr = 0.001) | 0.133(ANN lr = 0.5) |
| Incheon | Spring | 12.714(ANN lr = 0.3) | 17.671(ANN lr = 0.1) | 9.974(ANN lr = 0.1) | 0.098(ANN lr = 0.5) |
| | Summer | 9.067(LSTM-PC lr = 0.003) | 9.902(LSTM-PC lr = 0.007) | 7.132(LSTM-PC lr = 0.003) | 0.059(LSTM-PC lr = 0.003) |
| | Autumn | 10.798(LSTM lr = 0.001) | 14.206(LSTM-PC lr = 0.009) | 8.463(LSTM lr = 0.001) | 0.081(LSTM lr = 0.001) |
| | Winter | 9.66(LSTM-PC lr = 0.001) | 11.946(LSTM-PC lr = 0.001) | 7.572(LSTM-PC lr = 0.001) | 0.077(LSTM-PC lr = 0.001) |
| Gwangju | Spring | 14.076(LSTM lr = 0.001) | 19.601(LSTM lr = 0.001) | 11.404(LSTM lr = 0.001) | 0.111(LSTM lr = 0.003) |
| | Summer | 7.193(ANN lr = 0.1) | 6.401(ANN lr = 0.1) | 5.333(ANN lr = 0.1) | 0.044(ANN lr = 0.1) |
| | Autumn | 9.013(LSTM lr = 0.003) | 10.56(LSTM lr = 0.003) | 7.045(LSTM lr = 0.003) | 0.061(LSTM lr = 0.003) |
| | Winter | 12.555(LSTM-PC lr = 0.001) | 14.88(ANN lr = 0.3) | 9.785(ANN lr = 0.3) | 0.096(LSTM-PC lr = 0.001) |
| Pohang | Spring | 14.421(LSTM lr = 0.001) | 24.822(LSTM-PC lr = 0.003) | 12.089(LSTM lr = 0.001) | 0.124(LSTM lr = 0.001) |
| | Summer | 7.81(ANN lr = 0.7) | 8.028(ANN lr = 0.1) | 6.365(LSTM lr = 0.007) | 0.049(ANN lr = 0.7) |
| | Autumn | 9.349(ANN lr = 0.3) | 11.208(ANN lr = 0.3) | 7.621(ANN lr = 0.3) | 0.065(ANN lr = 0.3) |
| | Winter | 12.935(ANN lr = 0.3) | 17.59(ANN lr = 0.3) | 9.808(ANN lr = 0.3) | 0.112(ANN lr = 0.3) |
| Mokpo | Spring | 11.454(ANN lr = 0.5) | 15.142(ANN lr = 0.5) | 9.526(ANN lr = 0.5) | 0.082(ANN lr = 0.5) |
| | Summer | 5.732(LSTM lr = 0.005) | 5.263(LSTM lr = 0.005) | 4.296(LSTM lr = 0.005) | 0.036(LSTM lr = 0.005) |
| | Autumn | 6.951(ANN lr = 0.1) | 7.647(ANN lr = 0.1) | 5.393(ANN lr = 0.1) | 0.047(ANN lr = 0.1) |
| | Winter | 8.109(ANN lr = 0.1) | 8.971(ANN lr = 0.1) | 6.391(ANN lr = 0.1) | 0.058(ANN lr = 0.1) |
| Tongyeong | Spring | 10.427(LSTM-PC lr = 0.003) | 14.22(LSTM-PC lr = 0.003) | 8.55(LSTM-PC lr = 0.003) | 0.077(LSTM-PC lr = 0.003) |
| | Summer | 6.557(ANN lr = 0.1) | 6.227(LSTM-PC lr = 0.003) | 5.085(LSTM-PC lr = 0.003) | 0.039(ANN lr = 0.1) |
| | Autumn | 8.572(ANN lr = 0.1) | 9.732(LSTM lr = 0.005) | 6.729(LSTM lr = 0.005) | 0.059(ANN lr = 0.1) |
| | Winter | 12.672(ANN lr = 0.1) | 15.113(ANN lr = 0.1) | 9.415(ANN lr = 0.1) | 0.102(ANN lr = 0.5) |
| Jeonju | Spring | 12.046(LSTM lr = 0.003) | 14.774(LSTM-PC lr = 0.009) | 9.768(LSTM lr = 0.003) | 0.088(LSTM lr = 0.003) |
| | Summer | 6.751(ANN lr = 0.9) | 6.225(LSTM lr = 0.009) | 5.308(ANN lr = 0.9) | 0.039(ANN lr = 0.9) |
| | Autumn | 7.58(LSTM lr = 0.001) | 8.365(LSTM lr = 0.001) | 6.101(LSTM lr = 0.001) | 0.049(LSTM lr = 0.001) |
| | Winter | 8.965(ANN lr = 0.1) | 11.145(ANN lr = 0.1) | 7.015(ANN lr = 0.1) | 0.067(ANN lr = 0.1) |

As our result, Figs. 3 and 4 show, respectively, the lowest predicted values of MAE and Theil's-U statistic of five neural network models for all three kinds of epochs in four seasons of testing 3. The lowest predicted values of RMSE and MAPE of ten cities including all five neural netowrk models (the ANN, EML, DNN, LSTM, and LSTM-PC) are calculated for all three kinds of epochs in four seasons of ten cities in testing 4, as shown in Figs. 5 and 6. Tables 1 and 2 is illustrated the comparison of the RMSE, MAPE, MAE and Theil's-U statistic in four seasons of ten cities in testing 3 and testing 4, respectively.

From Fig. 3 and Table 1, the MAE of LSTM has the first lowest value of 0.647 in summer in Tongyeong (rank1), while the second one is 0.721 in summer in Mokpo and the third one is 0.967 in summer in Jeonju). The first lowest MAPE value of LSTM in summer in Tongyeong is 1.457 lower than 1.506 in summer in Mokpo (the second one) and 2.097 in summer in Jeonju (the third one), as seen from Table 1. From Fig. 5 and Table 2, the first RMSE of LSTM has a lowest value of 5.732 in summer in Mokpo, significantly rather than 6.557 in summer in Tongyeong (the seond one) and 6.557 in summer in Jeonju (the third one). The first lowest MAPE value of LSTM in summer in Mokpo is 5.263 lower than 6.225 in summer in Jeonju (the second one) and 6.227 in summer in Tongyeong (the third one), as depicted in Fig. 6.

Hence, we find that the RMSE of LSTM in temperature prediction has a lowest value of 5.732 at lr=0.005 in summer in Mokpo, while the lowest MAE value of LSTM is 0.647 at lr=0.01 in summer in Tongyeong.

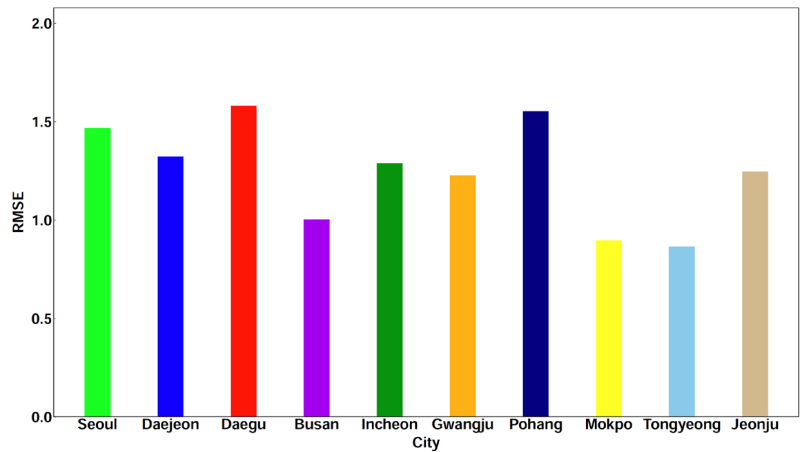

**Fig. 7:** The lowest RMSE of temperature prediction of ten cities for five NN models in four seasons in testing 1 (see TableA1 in Appendix A) and testing 3.

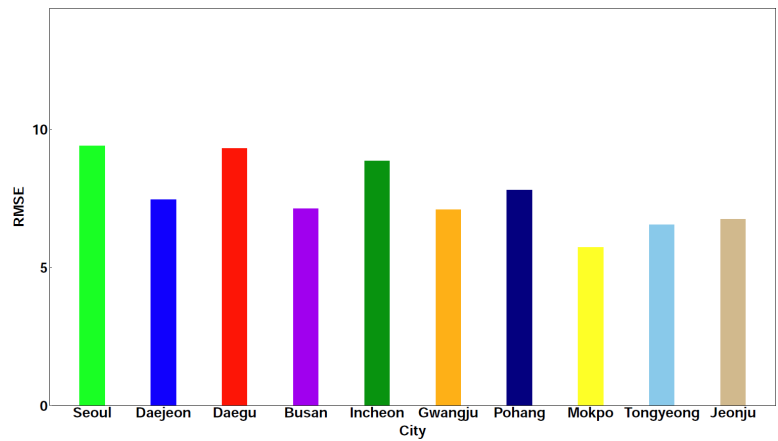

**Fig. 8:** The lowest RMSE of humidity prediction of ten cities for five NN models in four seasons in testing 2 (Table A2 in Appendix A) and testing 4.

**Table. 3:** The lowest RMSE values of temperature prediction, where lr denotes the learning rate.

| City | Season | Model | RMSE (unit: %) | Testing (Table) |
|---|---|---|---|---|
| Seoul | Summer | ANN (lr = 0.3) | 1.468 | 3 (1) |
| Incheon | Summer | LSTM-PC (lr = 0.009) | 1.288 | 3 (1) |
| Daejeon | Summer | LSTM-PC (lr = 0.001) | 1.322 | 3 (1) |
| Daegu | Summer | LSTM (lr = 0.001) | 1.58 | 1 (A1) |
| Busan | Summer | LSTM (lr = 0.001) | 1.003 | 1 (A1) |
| Pohang | Summer | LSTM (lr = 0.005) | 1.554 | 1 (A1) |
| Tongyeong | Summer | LSTM (lr = 0.005) | 0.866 | 3 (1) |
| Gwangju | Summer | LSTM-PC (lr = 0.001) | 1.227 | 1 (A1) |
| Jeonju | Summer | LSTM-PC (lr = 0.001) | 1.246 | 3 (1) |
| Mokpo | Summer | DNN (lr = 0.1) | 0.896 | 3 (1) |

**Table. 4:** Lowest RMSE values of humidity prediction, where lr denotes the learning rate.

| City | Season | Model | RMSE (unit: %) | Testing (Table) |
|---|---|---|---|---|
| Seoul | Summer | LSTM-PC (lr = 0.003) | 9.399 | 2 (A2) |
| Incheon | Summer | LSTM-PC (lr = 0.005) | 8.864 | 2 (A2) |
| Daejeon | Summer | ANN (lr = 0.1) | 7.46 | 2 (A2) |
| Daegu | Summer | ANN (lr = 0.1) | 9.307 | 2 (A2) |
| Busan | Summer | ANN (lr = 0.1) | 7.139 | 4 (2) |
| Pohang | Summer | ANN (lr = 0.7) | 7.81 | 4 (2) |
| Tongyeong | Summer | LSTM (lr = 0.009) | 6.549 | 2 (A2) |
| Gwangju | Summer | ANN (lr = 0.7) | 7.093 | 2 (A2) |
| Jeonju | Summer | ANN (lr = 0.9) | 6.751 | 4 (2) |
| Mokpo | Summer | LSTM (lr = 0.005) | 5.732 | 4 (2) |

In this subsection, we compare the respective predictive values of temperature and humidity from five neural network models in testings1, 2, 3, and 4. Fig. 7 (Fig. 8) shows the lowest RMSE of temperature prediction and humidity prediction of ten cities for the ANN, DNN, LSTM, LSTM-PC, and ELM in four seasons.

From Fig. 7 and Table 3, in temperature prediction, the RMSE of LSTM has the first lowest value of 0.866 in summer in Tongyeong, rather than 1.003 in summer in Busan (the second one) and 1.227 in summer in Gawngju (the third one). From Fig. 8 and Table 4, in humidity prediction, the RMSE of LSTM has the first lowest value of 5.732 in summer in Mokpo, rather than 6.549 in summer in Tongyeong (the second one) and 6.751 in summer in Jeonju (the third one).

Particularly, from the computer-simulation in order to predict the temperature in spring, the RMSE of the ANN in Tongyeong shows the lowest value for 7500 training epochs in testing 3 (average temperature predicted in the input layer with six input nodes). In summer, The RMSE of the LSTM in testing 3 has the lowest value in Tongyeong for 5000 training epochs. In the autumn, The LSTM-PC in testing 1 (average temperature predicted in the input layer with four input nodes) has the lowest value in Busan for 5000 training epoch. In winter, the LSTM-PC in testing 3 shows the lowest error in 5,000 training epoch of Daegu. In the temperature prediction, when using the LSTM model in testing 3 in Tongyeong in the summer among the four seasons, we find that the lowest value of RMSE is 0.866. In the simulation to predict the humidity in spring, the LSTM-PC in Tongyeong for 7500 training epochs has the lowest RMSE in testing 4 (average humidity predicted in the input layer with six input nodes). The LSTM in testing 4 has the lowest value in 2500 training epochs of Mokpo in the summer, while the RMSE of the LSTM in testing 2 (average humidity predicted in the input layer with four input nodes) in the autumn has the lowest value for 7500 training epochs in Mokpo. In winter, the RMSE of the ANN has lowest value in testing 4 (average humidity predicted in the input layer with six input nodes) trained 7500 epochs in Mokpo. In the humidity prediction, when using in the summer in Mokpo, the RMSE of LSTM model is shown the lowst value with 5.732. In both the average temperature and humidity predictions, the RMSEs are the lowest in summer.

## 4. Conclusion

In this paper, we have developed, trained, and tested the daily time series forecasting models for average temperature and average humidity in 10 major cities (Seoul, Daejeon, Daegu, Busan, Incheon, Gwangju, Pohang, Mokpo, Tongyeong, and Jeonju) using the NN models. We also have simulated the predictive accuracy in the ANN, DNN, ELM, LSTM, and LSTM-PC models. We introduce two testings: that is, input six nodes $T_{t-2}$, $T_{t-1}$, $T_t$, $H_{t-2}$, $H_{t-1}$, $H_t$. We $T_{t+1}$ and $H_{t+1}$ days of average temperature (testing 3) and average humidity (testing 4). As other cases, we performed the computer-simulation in Appendix A [103], for input four nodes $T_{t-1}$, $T_t$, $H_{t-1}$, $H_t$ days of temperature (testing 1) and humidity (testing 2). Hence, we can compare the model performances for input structures and lead times. In testings 1-4, the five learning rates for the ANN and the DNN are set to 0.1, 0.3, 0.5, 0.7, and 0.9, while those for LSTM and LSTM-PC are set to 0.001, 0.003, 0.005, 0.007, and 0.009 for 2500, 5000, and 7500 training epochs. The predicted values of the ELM are obtained by averaging the results trained 2500, 5000, and 7500 epochs. From the result of outputs, the root mean squared error (RMSE), mean absolute percentage error (MAPE), mean absolute error (MAE), Theil-U statistics are simulated for performance evaluation, and we compare each other after manipulating five NN models.

The two cases from our numerical calculation are obtained as follows: (1) In testing 3, the RMSE value is 1.563 for 7500 training epochs of the ANN (lr = 0.1) in spring in Tongyeong, and the LSTM (lr = 0.005) has an RMSE of 0.866 for 5000 training epochs in summer in Tongyeong. The RMSE value is 1.806 for 5000 training epochs of the LSTM-PC (lr = 0.001) in autumn in Pohang, while that is 2.081 for the 7500 traning epoch of The LSTM-PC (lr = 0.009) in winter in Daegu. The LSTM in summer and the LSTM-PC in autumn and winter show good performances, and as in testing 1, the LSTM series also show good performances. (2) In testing 4, the RMSE value is 10.427 for the 7500 training epochs of the LSTM-PC (lr = 0.003) in spring in Tongyeong. The RMSE value of LSTM (lr = 0.005) is 5.732 for 2500 training epochs in summer in Mokpo. The RMSE value is 6.951 for 2500 training epochs of the ANN (lr = 0.1) in summer in Mokpo, while that (lr = 0.1) is 8.109 for 7500 traning epochs in winter. In this case, The LSTM value outperforms any values of other models in summer in Mokpo. In addition, in testing A1, the RMSE value is 1.583 for the 7500 training epochs of the ANN (learning rate lr = 0.1) in spring in Tongyeong. The RMSE value of the LSTM-PC (lr = 0.003) in summer in Tongyeong is 0.878 for 2500 traning epochs, while that for the LSTM-PC (lr = 0.005) is 1.72 for the LSTM-PC (lr = 0.005) for 5000 traning epochs in autumn in Busan. The RMSE value of the LSTM-PC (lr = 0.007) is 2.078 for 5000 training epochs in winter in Daegu. Particularly, when the LSTM (lr = 0.003) is trained 2500 training epochs in summer in Tongyeong, the RMSE has the lowest value with 0.878. In testing A2, In spring, the RMSE value of the LSTM (lr = 0.001) is 10.609 for the 7500 training epochs in spring in Tongyeong. The RMSE value of the ANN (lr = 0.1) is 5.839 for 2500 training epochs in summer in Mokpo. The RMSE of the LSTM (lr = 0.005) was 6.891 for 7500 traning epochs in autumn in Mokpo, The RMSE value is 8.16 for 2500 training epochs of the ANN (lr = 0.1) in winter in Mokpo. When the ANN (lr = 0.1) is trained 2500 times in the summer in Mokpo, the RMSE has the lowest value with 5.839. Consequently, we find that the RMSE of LSTM in temperature prediction has a lowest value of 0.866 at learning rate lr=0.005 in summer in Tongyeung, while the lowest RMSE value of LSTM in humidity prediction is 5.732 at learning rate lr=0.005 in summer in Tongyeong.

From numerical results, the difference between the actual value and the predicted value of humidity is relatively greater than the average temperature, and the reason for this is that the actual value of humidity is more chaotic than that of temperature as shown in the previous paper [103,104]. The reason is that the meteorological factor in inland cities may be given less error than that of coastal cities. We also consider the data of inland cities would have inherently more chaotic and non-linear time series. Our result provides the evidence that the LSTM is an effective method of predicting one meteorological factor (temperature) rather than the DNN. Our result for predictive temperature is approximately consistent to that of Chen et al. [87], which was obtained from Chinese stock data via the DNN model. Wei also obtained a good result that the RMSE has about 0.05% prediction value using data selected from 20 stock datasets [99], and the LSTM [100] for future stock prices with previous two week's data as input has the average RMSE value less than 0.05.

We will conduct a study to further improve the accuracy of the meteorological element prediction model by applying a learning method that applies optimization algorithms such as genetic algorithm and particle swarm optimization to other types of neural network models such as and backpropagation algorithms [105,106]. There exists complicatedly and rebelliously correlated relation between several meteorological factors such temperature, wind velocity, humidity, surface hydrology, heat transfer, solar radiation, surface hydrology, land subsidence, and so on. The research in future can treat and apply complex network theory to input variables of the meteorological data, and the LSTM and LSTM-PC models can promote and develop the predictive performance. We consider in the future that the LSTM and the LSTM-PC have excellent results of reducing error if the learning rate value and the number of epochs are adequately tried and regulated for a specific testing problem

## Appendix A : Testings 1 and 2

As the result of Ref. [103], testing 1 has the four nodes $T_{t-1}$, $T_t$, $H_{t-1}$, $H_t$, in the input layer and the one output node $T_{t+1}$ in output layer. Testing 2 has the four input nodes $T_{t-1}$, $T_t$, $H_{t-1}$, $H_t$, and the one output node $H_{t+1}$. It is not known beforehand what values of learning rates are appropriate. However, we select the five learning rate lr=0.1, 0.2, 0.3, 0.4, and 0.5 for the ANN and the DNN, while the learning rate values for LSTM and LSTM-PC are lr=0.001, 0.003, 0.005, 0.007, and 0.009, for different training sizes over three runs, 2500, 5000, and 7500 epochs. Particularly, the predicted accuracies of ELM are also obtained by averaging the results over 2500, 5000, and 7500 epochs. Table A1 and Table A2 (the same as Table 1 and 2 in Ref. [103]) are, respectively, the result of the computer-simulation performed for testing 1 and 2.

**Table A1.** Values of RMSE, MAPE, MAE and Theil's-U (unit: %) of ten metropolitan cities in four seasons in testing 1, where lr and L-PC denote the learning rate and the LSTM-PC, respectively.

| City | Season | RMSE | MAPE | MAE | Theil's-U ($\times 10^{-3}$) |
|---|---|---|---|---|---|
| Seoul | Spring | 2.187(LSTM, lr=0.001) | 0.61 (DNN, lr=0.1) | 1.747 (DNN, lr=0.1) | 3.823(LSTM, lr=0.001) |
| | Summer | 1.469(LSTM, lr=0.001) | 0.392(LSTM, lr=0.001) | 1.169(LSTM, lr=0.001) | 2.462(LSTM, lr=0.001) |
| | Autumn | 2.126 (L-PC, lr=0.005) | 0.392 (L-PC, lr=0.009) | 1.575 (L-PC, lr=0.005) | 3.686 (L-PC, lr=0.005) |
| | Winter | 2.832 (ANN, lr=0.5) | 0.552 (L-PC, lr=0.005) | 2.046 (DNN, lr=0.3) | 5.153 (ANN, lr=0.5) |
| Daejeon | Spring | 2.170 (ANN, lr=0.1) | 0.591 (DNN, lr=0.1) | 1.694 (DNN, lr=0.1) | 3.785 (ANN, lr=0.1) |
| | Summer | 1.339(LSTM, lr=0.001) | 0.333(LSTM, lr=0.007) | 0.992(LSTM, lr=0.007) | 2.244(LSTM, lr=0.001) |
| | Autumn | 2.227 (L-PC, lr=0.007) | 0.580 (L-PC, lr=0.007) | 1.660 (L-PC, lr=0.007) | 3.855 (L-PC, lr=0.007) |
| | Winter | 2.465 (ANN, lr=0.5) | 0.692 (ANN, lr=0.5) | 1.909 (ANN, lr=0.5) | 4.469 (ANN, lr=0.5) |
| Daegu | Spring | 2.491 (L-PC, lr=0.007) | 0.677 (L-PC, lr=0.009) | 1.947 (L-PC, lr=0.009) | 4.328 (L-PC, lr=0.007) |

|  | Summer | 1.580(LSTM, lr=0.001) | 0.407(LSTM, lr=0.001) | 1.209(LSTM, lr=0.001) | 2.646(LSTM, lr=0.001) |
|  | Autumn | 1.852(LSTM, lr=0.009) | 0.491 (L-PC, lr=0.007) | 1.412 (L-PC, lr=0.007) | 3.202(LSTM, lr=0.009) |
|  | Winter | 2.046 (ANN, lr=0.3) | 0.545 (ANN, lr=0.3) | 1.509 (ANN, lr=0.3) | 3.694 (ANN, lr=0.3) |
| Busan | Spring | 1.936 (DNN, lr=0.1) | 0.522 (ANN, lr=0.3) | 1.499 (ANN, lr=0.3) | 3.366 (DNN, lr=0.1) |
|  | Summer | 1.003(LSTM, lr=0.001) | 0.266(LSTM, lr=0.001) | 0.789(LSTM, lr=0.001) | 1.686(LSTM, lr=0.001) |
|  | Autumn | 1.720(LSTM, lr=0.005) | 0.448 (L-PC, lr=0.005) | 1.298(LSTM, lr=0.005) | 2.955 (L-PC, lr=0.005) |
|  | Winter | 2.375 (DNN, lr=0.3) | 0.640 (DNN, lr=0.1) | 1.793 (DNN, lr=0.1) | 4.241 (DNN, lr=0.3) |
| Incheon | Spring | 1.795 (ANN, lr=0.1) | 0.493 (DNN, lr=0.1) | 1.405 (ANN, lr=0.1) | 3.149 (ANN, lr=0.1) |
|  | Summer | 1.294(LSTM, lr=0.009) | 0.334(LSTM, lr=0.009) | 0.997(LSTM, lr=0.009) | 2.173(LSTM, lr=0.009) |
|  | Autumn | 2.163(LSTM, lr=0.009) | 0.569 (ANN, lr=0.3) | 1.628 (ANN, lr=0.3) | 3.746 (ANN, lr=0.3) |
|  | Winter | 2.612 (L-PC, lr=0.005) | 0.726 (DNN, lr=0.5) | 1.991 (DNN, lr=0.5) | 4.751 (L-PC, lr=0.005) |
| Gwangju | Spring | 2.108 (L-PC, lr=0.007) | 0.565 (ANN, lr=0.1) | 1.617 (ANN, lr=0.1) | 3.674 (L-PC, lr=0.007) |
|  | Summer | 1.227 (L-PC, lr=0.001) | 0.314 (L-PC, lr=0.001) | 0.934 (L-PC, lr=0.001) | 2.058 (L-PC, lr=0.001) |
|  | Autumn | 2.086(LSTM, lr=0.005) | 0.518 (L-PC, lr=0.007) | 1.487 (L-PC, lr=0.007) | 3.599(LSTM, lr=0.005) |
|  | Winter | 2.418(LSTM, lr=0.001) | 0.633 (ANN, lr=0.3) | 1.762 (ANN, lr=0.3) | 4.354(LSTM, lr=0.001) |
| Pohang | Spring | 2.754(LSTM, lr=0.001) | 0.745(LSTM, lr=0.009) | 2.147(LSTM, lr=0.009) | 4.783(LSTM, lr=0.001) |
|  | Summer | 1.554(LSTM, lr=0.005) | 0.436(LSTM, lr=0.001) | 1.299(LSTM, lr=0.001) | 2.606(LSTM, lr=0.005) |
|  | Autumn | 1.801 (ANN, lr=0.1) | 0.482 (L-PC, lr=0.007) | 1.393 (L-PC, lr=0.007) | 3.101 (ANN, lr=0.1) |
|  | Winter | 2.249 (DNN, lr=0.1) | 0.598 (DNN, lr=0.1) | 1.666 (DNN, lr=0.1) | 4.041 (DNN, lr=0.1) |
| Mokpo | Spring | 1.81 (LSTM, lr=0.001) | 0.511(LSTM, lr=0.001) | 1.458(LSTM, lr=0.001) | 3.163(LSTM, lr=0.001) |
|  | Summer | 0.916 (L-PC, lr=0.007) | 0.243 (L-PC, lr=0.007) | 0.722 (L-PC, lr=0.007) | 1.539 (L-PC, lr=0.007) |
|  | Autumn | 1.980 (ANN, lr=0.1) | 0.492 (L-PC, lr=0.005) | 1.413 (L-PC, lr=0.005) | 3.416 (ANN, lr=0.1) |
|  | Winter | 2.310 (ANN, lr=0.5) | 0.611 (ANN, lr=0.5) | 1.695 (ANN, lr=0.5) | 4.166 (ANN, lr=0.5) |
| Tongyeong | Spring | 1.583 (ANN, lr=0.1) | 0.439(LSTM, lr=0.005) | 1.26 (LSTM, lr=0.005) | 2.758 (ANN, lr=0.1) |
|  | Summer | 0.878 (L-PC, lr=0.003) | 0.226 (L-PC, lr=0.003) | 0.671 (L-PC, lr=0.003) | 1.478 (L-PC, lr=0.003) |
|  | Autumn | 1.783 (L-PC, lr=0.005) | 0.463 (L-PC, lr=0.005) | 1.336 (L-PC, lr=0.005) | 3.063 (L-PC, lr=0.005) |
|  | Winter | 2.144 (ANN, lr=0.3) | 0.585 (ANN, lr=0.3) | 1.635 (ANN, lr=0.3) | 3.840 (ANN, lr=0.3) |
| Jeonju | Spring | 2.281 (L-PC, lr=0.005) | 0.622 (DNN, lr=0.1) | 1.779 (DNN, lr=0.1) | 3.983 (L-PC, lr=0.005) |
|  | Summer | 1.246 (L-PC, lr=0.001) | 0.324 (L-PC, lr=0.003) | 0.961 (L-PC, lr=0.003) | 2.090 (L-PC, lr=0.001) |
|  | Autumn | 2.220(LSTM, lr=0.009) | 0.584 (L-PC, lr=0.007) | 1.675 (L-PC, lr=0.007) | 3.837(LSTM, lr=0.009) |
|  | Winter | 2.601 (ANN, lr=0.3 ) | 0.716 (ANN, lr=0.3) | 1.984 (ANN, lr=0.3) | 4.699 (ANN, lr=0.3) |

**Table A2.** Values of RMSE, MAPE, MAE and Theil's-U (unit: %) of ten metropolitan cities in four seasons in testing 2, where lr and L-PC denote the learning rate and the LSTM-PC, respectively.

| City | Season | RMSE | MAPE | MAE | Theil's-U ($\times 10^{-3}$) |
|---|---|---|---|---|---|
| Seoul | Spring | 13.302 (L-PC, lr=0.001) | 22.944 (L-PC, lr=0.009) | 10.7 (L-PC, lr=0.009) | 0.126 (L-PC, lr=0.001) |
|  | Summer | 9.399 (L-PC, lr=0.003) | 11.087 (L-PC, lr=0.007) | 7.226 (LSTM, lr=0.009) | 0.071 (L-PC, lr=0.003) |
|  | Autumn | 11.313 (L-PC, lr=0.001) | 14.0 (LSTM, lr=0.007) | 8.055 (LSTM, lr=0.007) | 0.091 (L-PC, lr=0.001) |
|  | Winter | 11.054 (L-PC, lr=0.001) | 14.915 (L-PC, lr=0.001) | 8.531 (L-PC, lr=0.001) | 0.095 (L-PC, lr=0.001) |
| Daejeon | Spring | 11.468 (L-PC, lr=0.001) | 17.095 (L-PC, lr=0.001) | 9.640 (L-PC, lr=0.001) | 0.094 (L-PC, lr=0.001) |
|  | Summer | 7.460 (ANN, lr=0.1) | 7.433(LSTM, lr=0.009) | 5.737 (ANN, lr=0.1) | 0.048 (ANN, lr=0.1) |
|  | Autumn | 7.827 (L-PC, lr=0.001) | 7.990 (L-PC, lr=0.005) | 5.809 ((L-PC, lr=0.001) | 0.051 (L-PC, lr=0.001) |
|  | Winter | 10.403 (LSTM, lr=0.001) | 11.352 (L-PC, lr=0.001) | 7.862 (L-PC, lr=0.001) | 0.074 (L-PC, lr=0.007) |
| Daegu | Spring | 12.798 (LSTM, lr=0.001) | 21.951 (L-PC, lr=0.003) | 10.176 (L-PC, lr=0.001) | 0.121(LSTM, lr=0.001) |
|  | Summer | 9.307 (ANN, lr=0.1) | 9.392 (L-PC, lr=0.007) | 6.996 (L-PC, lr=0.007) | 0.064 (ANN, lr=0.1) |
|  | Autumn | 9.614 (L-PC, lr=0.003) | 10.853 (L-PC, lr=0.003) | 7.504 (LSTM, lr=0.001) | 0.067(LSTM, lr=0.001) |
|  | Winter | 13.404(LSTM, lr=0.001) | 16.484 (L-PC, lr=0.001) | 9.594 (L-PC, lr=0.001) | 0.114 (L-PC, lr=0.005) |
| Busan | Spring | 12.085(LSTM, lr=0.001) | 19.212(LSTM, lr=0.001) | 9.893 (LSTM, lr=0.001) | 0.101 (L-PC, lr=0.003) |
|  | Summer | 7.166 (ANN, lr=0.1) | 7.147 (ANN, lr=0.1) | 5.708 (ANN, lr=0.1) | 0.045 (ANN, lr=0.1) |
|  | Autumn | 10.032 (L-PC, lr=0.009) | 12.081(LSTM, lr=0.007) | 7.744(LSTM, lr=0.007) | 0.072(LSTM, lr=0.005) |
|  | Winter | 14.173 (L-PC, lr=0.001) | 18.804 (L-PC, lr=0.003) | 10.081 (L-PC, lr=0.001) | 0.131 (ANN, lr=0.7) |
| Incheon | Spring | 12.629 (ANN, lr=0.5) | 17.822 (ANN, lr=0.3) | 9.989 (ANN, lr=0.3) | 0.097 (ANN, lr=0.5) |
|  | Summer | 8.864 (L-PC, lr=0.005) | 9.755 (L-PC, lr=0.005) | 6.969 (L-PC, lr=0.005) | 0.057 (L-PC, lr=0.005) |
|  | Autumn | 10.941 (L-PC, lr=0.005)) | 14.308 (L-PC, lr=0.005) | 8.640 (L-PC, lr=0.005) | 0.083 (L-PC, lr=0.005) |
|  | Winter | 9.832 (L-PC, lr=0.009) | 11.892 (L-PC, lr=0.003) | 7.585 (L-PC, lr=0.003) | 0.078 (L-PC, lr=0.009) |
| Gwangju | Spring | 13.862 (LSTM, lr=0.001) | 19.23 (LSTM, lr=0.001) | 11.248 (LSTM, lr=0.001) | 0.109(LSTM, lr=0.001) |
|  | Summer | 7.093 (ANN, lr=0.7) | 6.659 (L-PC, lr=0.007) | 5.530 (L-PC, lr=0.007) | 0.044 (ANN, lr=0.7) |
|  | Autumn | 8.895(LSTM, lr=0.003) | 10.391(LSTM, lr=0.003) | 6.945(LSTM, lr=0.003) | 0.059(LSTM, lr=0.003) |
|  | Winter | 12.703 (L-PC, lr=0.005) | 15.463 (ANN, lr=0.5) | 10.008 (L-PC, lr=0.005) | 0.096 (L-PC, lr=0.009) |
| Pohang | Spring | 14.322(LSTM, lr=0.001) | 24.717 ((L-PC, lr=0.003) | 11.793(LSTM, lr=0.001) | 0.123(LSTM, lr=0.001) |
|  | Summer | 7.990 (ANN, lr=0.7) | 8.122(LSTM, lr=0.001) | 6.326(LSTM, lr=0.001) | 0.051 (ANN, lr=0.9) |
|  | Autumn | 9.408 (ANN, lr=0.3) | 11.043 (ANN, lr=0.3) | 7.598 (ANN, lr=0.3) | 0.065 (ANN, lr=0.3) |
|  | Winter | 12.832 (ANN, lr=0.5) | 17.431(LSTM, lr=0.007) | 9.866 (ANN, lr=0.5) | 0.109 (ANN, lr=0.5) |
| Mokpo | Spring | 11.435 (ANN, lr=0.7) | 15.024 (ANN, lr=0.7) | 9.502 ((L-PC, lr=0.007) | 0.082 (ANN, lr=0.7) |
| | Summer | 5.839 (ANN, lr=0.1) | 5.525(LSTM, lr=0.007) | 4.451(LSTM, lr=0.007) | 0.036 (ANN, lr=0.1) |
| | Autumn | 6.891(LSTM, lr=0.005) | 7.702 (ANN, lr=0.3) | 5.494 (ANN, lr=0.1) | 0.046(LSTM, lr=0.003) |
| | Winter | 8.160 (ANN, lr=0.1) | 9.248 (ANN, lr=0.1) | 6.540 (ANN, lr=0.1) | 0.058 (ANN, lr=0.1) |
| Tongyeong | Spring | 10.479(LSTM, lr=0.001) | 13.926 (L-PC, lr=0.003) | 8.613 (L-PC, lr=0.003) | 0.076(LSTM, lr=0.001) |
| | Summer | 6.549(LSTM, lr=0.009) | 6.166 (LSTM, lr=0.009) | 5.002(LSTM, lr=0.009) | 0.039(LSTM, lr=0.005) |
| | Autumn | 8.630 (ANN, lr=0.1) | 9.729 (LSTM, lr=0.007) | 6.747 (L-PC, lr=0.003) | 0.059 (ANN, lr=0.1) |
| | Winter | 12.371(LSTM, lr=0.001) | 14.826 (L-PC, lr=0.003) | 9.150 (L-PC, lr=0.001) | 0.101 (ANN, lr=0.7) |
| Jeonju | Spring | 12.011 (L-PC, lr=0.003) | 14.698(LSTM, lr=0.001) | 9.658 (LSTM, lr=0.001) | 0.088 (L-PC, lr=0.003) |
| | Summer | 7.027 (ANN, lr=0.9) | 6.351(LSTM, lr=0.007) | 5.412 (L-PC, lr=0.005) | 0.041 (ANN, lr=0.9) |
| | Autumn | 7.386 (L-PC, lr=0.001) | 8.178 (L-PC, lr=0.001) | 5.980 (L-PC, lr=0.001) | 0.047 (L-PC, lr=0.001) |
| | Winter | 8.899 (ANN, lr=0.1) | 11.033 (ANN, lr=0.1) | 6.992 (ANN, lr=0.1) | 0.066 (ANN, lr=0.1) |